\documentclass[11pt, a4paper]{article}

\usepackage[utf8]{inputenc}
\usepackage[T1]{fontenc}
\usepackage{geometry}
\usepackage{amsmath, amssymb, amsfonts}
\usepackage{graphicx}
\usepackage{booktabs} % 漂亮的表格
\usepackage{multirow} % 表格合并行
\usepackage{hyperref} % 超链接和引用跳转
\usepackage{authblk}  % 作者单位格式
\usepackage{url}
\usepackage{multibib}
\newcites{supp}{Supplementary References}
\graphicspath{{./figs/}}

\title{BoxMind: Closed-loop AI strategy optimization for elite boxing validated in the 2024 Olympics}

\author[1]{Kaiwen Wang\thanks{These authors contributed equally.}}
\author[1]{Kaili Zheng\textsuperscript{*}}
\author[2]{Rongrong Deng\textsuperscript{*}}
\author[2]{Qingmin Fan}
\author[1]{Milin Zhang}
\author[1]{Zongrui Li}
\author[1]{Xuesi Zhou}
\author[3]{Bo Han}
\author[3]{Liren Chen}
\author[1]{Chenyi Guo\thanks{Corresponding authors.}}
\author[1,4,5]{Ji Wu\textsuperscript{\textdagger}}

\affil[1]{Department of Electronic Engineering, Tsinghua University}
\affil[2]{Beijing Sport University}
\affil[3]{Chinese Boxing Federation}
\affil[4]{College of AI, Tsinghua University}
\affil[5]{Beijing National Research Center for Information Science and Technology}

\date{}

\begin{document}

\maketitle

% 摘要 
\begin{abstract}
Competitive sports require sophisticated tactical analysis, yet combat disciplines like boxing remain underdeveloped in AI-driven analytics due to the complexity of action dynamics and the lack of structured tactical representations. To address this, we present BoxMind, a closed-loop AI expert system validated in elite boxing competition. By defining atomic punch events with precise temporal boundaries and spatial and technical attributes, we parse match footage into 18 hierarchical technical-tactical indicators. We then propose a graph-based predictive model that fuses these explicit technical-tactical profiles with learnable, time-variant latent embeddings to capture the dynamics of boxer matchups. Modeling match outcome as a differentiable function of technical-tactical indicators, we turn winning probability gradients into executable tactical adjustments. Experiments show that the outcome prediction model achieves state-of-the-art performance, with 69.8\% accuracy on BoxerGraph test set and 87.5\% on Olympic matches. Using this predictive model as a foundation, the system generates strategic recommendations that demonstrate proficiency comparable to human experts. BoxMind is validated through a closed-loop deployment during the 2024 Paris Olympics, directly contributing to the Chinese National Team’s historic achievement of three gold and two silver medals. BoxMind establishes a replicable paradigm for transforming unstructured video data into strategic intelligence, bridging the gap between computer vision and decision support in competitive sports. Code and data is available at: https://github.com/gouba2333/BoxingWeb.
\end{abstract}

\section{Introduction}
Success in competitive sports relies on athletes’ ability to balance self-improvement with opponent-specific adaptations \cite{farrow2017development,schinke2012adaptation}. To dominate, athletes must not only train rigorously but also analyze opponents’ tactical patterns and devise countermeasures \cite{guidetti2002physiological,tyshchenko2023neurological}. This emphasis on tactical analysis has driven the adoption of quantitative tactical perception in competitive sports. By quantifying technical and tactical indicators, such as the proportions and success rates of different techniques, athletes and coaches can objectively map opponents’ strengths and vulnerabilities \cite{zadorozhna2020tactical,hatmaker2004boxing}.

Boxing, as a highly tactically demanding discipline, exemplifies this need for advanced analysis. A boxer’s performance constitutes a high-dimensional interplay of spatial control, stylistic adaptations, and rhythm management \cite{kapo2021winning,ashker2011technical,pic2021professional,davis2013amateur,slimani2017performance}. However, the methodologies for capturing these dynamics have remained traditional. Unlike many data-rich sports, boxing analysis still relies on manual video review and expert annotation \cite{devesa2020methodological,thomson2016technical}. This process is not only labor-intensive and subjective but also limits the consistency and scalability of tactical insights. 

Over the past decade, while AI has transformed analytics in team sports like football and basketball—enabling automated player tracking, event recognition, and granular tactical modeling \cite{acikmese2017towards,zhong2018study,horvat2019data,ofoghi2013supporting,puchun2016application,giancola2018soccernet,wu2022application,gang2022dynamic,wang2024tacticai}—its application in combat sports remains underdeveloped. Current efforts in boxing are largely confined to basic action classification \cite{manoharan2023punch,kasiri2017fine,stefanski2022detecting,stefanski2023classification,stefanski2024boxing,lai2024facts,baghel2024efficient} and struggle to extract sophisticated technical-tactical indicators due to the rapid and complex nature of combat dynamics. A critical gap lies between visual perception (identifying what happened) and high-level strategic reasoning (inferring how to win). 

While there is no universally accepted standard for strategy selection throughout a game, the final outcome of each match is always clear. Therefore, we propose taking the match outcome as the starting point to systematically analyze the causal relationships between the strategic patterns and the result. Here, match outcome prediction serves as the engine for strategy formulation rather than just a forecasting tool. Yet, existing predictive models have significant limitations. Traditional rating systems \cite{glickman1995glicko,elo1967proposed,coulom2008whole,boxrec2024} reduce multifaceted athletes to a single scalar value, obscuring stylistic nuances crucial to matchup dynamics. Conversely, simple indicator-based methods \cite{horvat2020use,chen2021hybrid,horvat2023data,prasetio2016predicting,yeung2023framework,huang2021use,yan2024artificial} rely on statistical averages but often neglect the context of opponent strength and athletes’ temporal evolution. This inability to model technical-tactical patterns and the dynamic interplay between latent capabilities fundamentally limits the transition from outcome forecasting to actionable strategy generation.

Ultimately, effective strategy should be derived from a robust understanding of causal factors driving match outcomes. However, current strategic methodologies remain disjointed: discrete event models \cite{wang2024tacticai} are ill-suited to the continuous flow of boxing, while aggregate statistical approaches often lack the granularity required for personalized advice \cite{yan2024artificial,chang2022constructing}. Consequently, a unified framework capable of automating tactical data extraction, modeling complex matchups for prediction, and recommending specific strategies has yet to be established.

In this paper, we propose BoxMind, a scalable AI expert system designed to bridge the gap between raw visual data and high-level strategic decision-making. The framework is built upon three core methodological contributions.

First, we construct a hierarchical technical-tactical indicator system. Moving beyond basic action recognition, we define atomic punch events characterized by precise temporal boundaries and spatial and technical attributes (range, target, technique, effect). These atomic events are aggregated to compute 18 distinct high-level indicators, providing a structured quantification of a boxer's style that serves as the interpretable representation for our system.

Second, we design a Graph-based Match Outcome Prediction Model. Instead of treating matches in isolation, we construct a boxer graph where each athlete is represented by a learnable, time-aware embedding. By fusing these latent competitive hierarchy embeddings with the explicit historical indicator profiles of both fighters, the model contextualizes a boxer's observable technical-tactical output within their global competitive standing, achieving superior predictive accuracy compared to traditional rating systems.

Third, we propose a Gradient-based Strategy Recommendation mechanism. By formulating the match outcome as a differentiable function of the input indicators, we can calculate the gradient of the winning probability with respect to specific tactical behaviors. This allows BoxMind to prescribe optimal strategic interventions—such as increasing specific punch combinations or altering engagement distances—to maximize the probability of winning against a specific opponent.

We validate BoxMind through a real-world closed-loop deployment during the 2024 Paris Olympics. The system is deployed to support the Chinese National Team, directly contributing to their historic achievement of three gold and two silver medals. As a representative case study, we detail the preparation of women’s 75kg champion Li Qian, demonstrating that AI-driven recommendations, when translated into training interventions, can tangibly enhance competitive performance.

\section{Methods}

\subsection{Hierarchical Technical-Tactical Indicator Extraction}

% Figure 1
\begin{figure}[htbp]
    \centering
    \includegraphics[width=\textwidth]{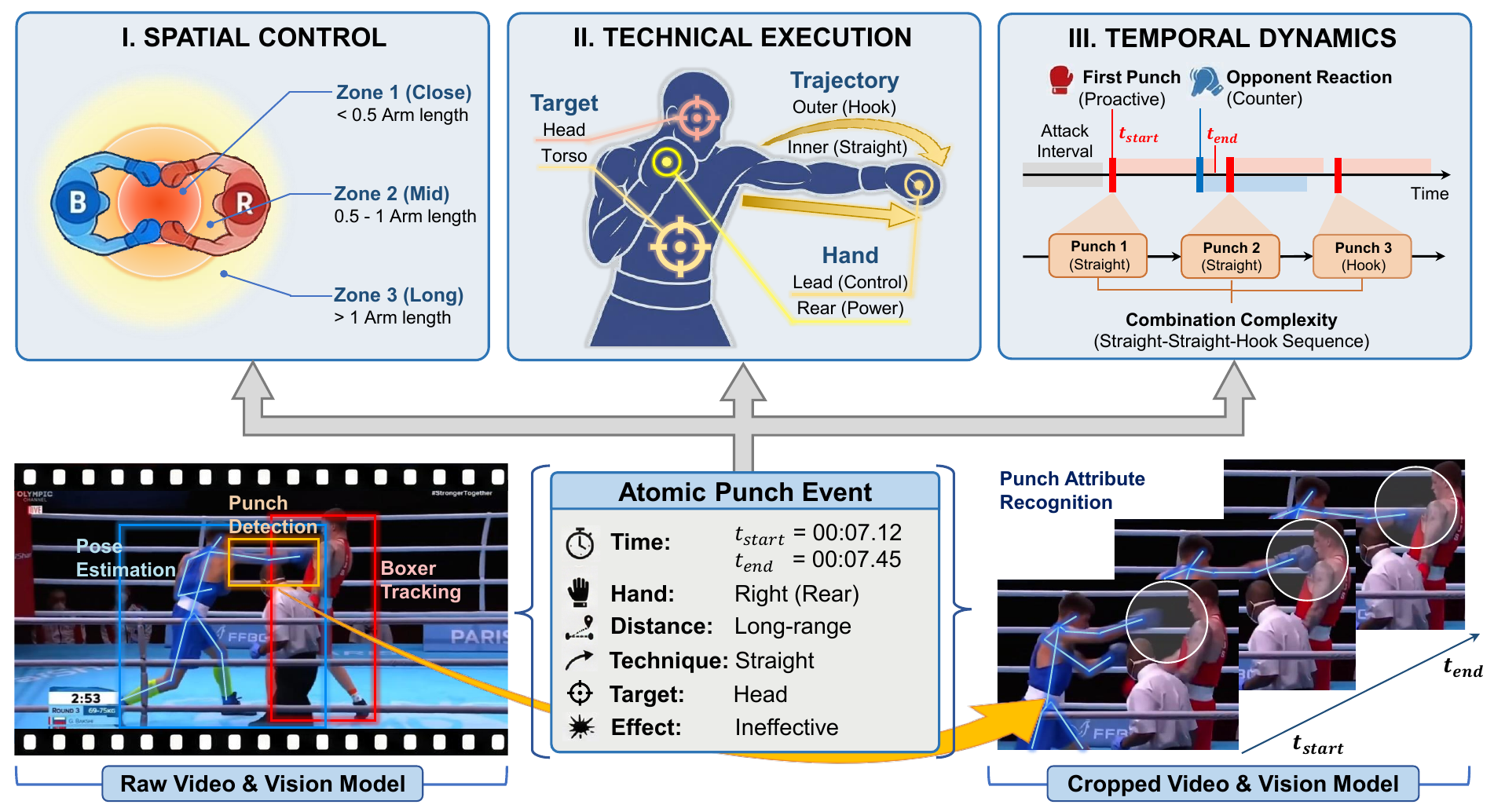} % 取消注释并替换为你实际的文件名
    \caption{From Atomic Vision to Hierarchical Technical-Tactical Intelligence. The framework transforms raw visual signals into structured knowledge.}
    \label{fig:framework}
\end{figure}

To bridge the gap between unstructured video data and expert-level decision support, we propose a hierarchical extraction framework (Fig. \ref{fig:framework}).This framework transforms raw visual signals into a structured knowledge representation through two stages: atomic punch event detection and technical-tactical indicator aggregation. 

\subsubsection{Atomic Punch Event Definition}
The foundation of our system is the discrete atomic punch event. Unlike continuous action recognition, we define a punch event e as a structured tuple containing precise temporal and spatial attributes:

\begin{equation}
    e = (t_{start}, t_{end}, a_{hand}, a_{dist}, a_{tech}, a_{target}, a_{eff})
    \label{eq:atomic_event}
\end{equation}

\noindent Where:
\begin{itemize}
    \item $t_{start}, t_{end}$: The precise start and end frames of the action;
    \item $a_{hand} \in \{Lead, Rear\}$: The hand delivering the strike;
    \item $a_{dist} \in \{Long, Mid, Close\}$: The spatial distance relative to the opponent;
    \item $a_{tech} \in \{Straight, Hook, Uppercut\}$: The trajectory type of the punch;
    \item $a_{target} \in \{Head, Torso\}$: The intended impact zone;
    \item $a_{eff} \in \{Effective, Ineffective\}$: Whether the punch landed cleanly with the knuckle area.
\end{itemize}

The detailed definition of each attribute is provided in the supplementary material.
To extract these attributes, we utilize a validated computer vision pipeline. This pipeline employs 4D-Humans \cite{goel2023humans} for pose estimation \cite{loper2023smpl}, a UV-map Enhanced method for boxer tracking, a TCN-based detection model \cite{lin2021learning,lea2017temporal} to localize temporal boundaries of punches, and an custom-designed Pose and Region Guided model \cite{carreira2017quo,jacobs1991adaptive} to classify the spatial and technical attributes. For details of the models, please refer to the supplementary material.

\subsubsection{Hierarchical Indicator Aggregation}

% Table 1 
\begin{table}[htbp]
    \centering
    \caption{The Hierarchical Technical-Tactical Indicator System. "Prop." and "No." denote "Proportion" and "Number (per minute)" respectively.}
    \label{tab:indicators}
    \resizebox{\textwidth}{!}{
    \begin{tabular}{lll}
        \toprule
        \textbf{Dimension} & \textbf{Category} & \textbf{Indicator}  \\
        \midrule
        \multirow{3}{*}{\textbf{I. Spatial Control}} & \multirow{3}{*}{A. Distance Management} & 1. Prop. of Close- \& Mid-Range Punches \\
         & & 2. No. of Effective Close- \& Mid-Range Punches \\
         & & 3. No. of Effective Long-Range Punches \\
        \midrule
        \multirow{10}{*}{\textbf{II. Technical Execution}} & \multirow{3}{*}{B. Hand Usage} & 4. Prop. of Lead Hand Punches  \\
         & & 5. No. of Effective Lead Hand Punches \\
         & & 6. No. of Effective Rear Hand Punches \\
         \cmidrule{2-3}
         & \multirow{3}{*}{C. Target Choice} & 7. Prop. of Punches Targeted at Torso  \\
         & & 8. No. of Effective Punches Targeted at Torso \\
         & & 9. No. of Effective Punches Targeted at Head \\
         \cmidrule{2-3}
         & \multirow{4}{*}{D. Trajectory Logic} & 10. Prop. of Straight Punches  \\
         & & 11. No. of Effective Straight Punches \\
         & & 12. Prop. of Mid- \& Long-Range Hook Punches \\
         & & 13. No. of Effective Mid- \& Long-Range Hook Punches \\
        \midrule
        \multirow{5}{*}{\textbf{III. Temporal Dynamics}} & \multirow{2}{*}{E. Attacking Rhythm} & 14. Prop. of Proactive Punches  \\
         & & 15. Prop. of Counter Punches \\
         \cmidrule{2-3}
         & \multirow{3}{*}{F. Combination Complexity} & 16. Prop. of Straight-Straight Combo  \\
         & & 17. Prop. of Hook Combo \\
         & & 18. Prop. of Uppercut Combo \\
        \bottomrule
    \end{tabular}
    }
\end{table}

Raw atomic events offer limited strategic value in isolation. To capture the stylistic signature of a boxer, we aggregate these events into 18 high-level technical-tactical indicators (Listed in Table \ref{tab:indicators}). These indicators are organized into a three-dimensional hierarchy: Spatial Control, Technical Execution, and Temporal Dynamics—which are widely recognized as core dimensions in boxing and combat sports performance analysis \cite{kapo2021winning,ashker2011technical,pic2021professional,davis2013amateur,slimani2017performance}.

Spatial Control quantifies where the boxer engages relative to the opponent. Within Distance Management, we calculate the volume and efficiency of punches across different ranges. We group close and mid-range punches due to their shared infighting characteristics, contrasting them with long-range punches that reflect out-boxing strategies. Consequently, a high proportion of close and mid-range punches indicates an infighting style, whereas dominance at long range implies an out-boxing strategy utilizing reach advantages \cite{kapo2021winning,slimani2017performance}.

Technical Execution analyzes the delivery mechanisms of attacks, consisting of Hand Usage, Target Choice and Trajectory Logic. Hand Usage distinguishes between pace control via the lead hand and power delivery via the rear hand. Target Choice examines the vertical distribution of attacks, where a higher proportion of torso strikes reveals strategies for level-changing and stamina depletion. Trajectory Logic differentiates between inner linear straight punches and outer angular hooks. Notably, frequent mid and long-range hooks reflect an intention to bypass a tight high-guard using arc trajectories.

Temporal Dynamics captures the timing and sequencing of attacks. Under Attacking Rhythm, we distinguish initiative by defining an attack interval as a pause exceeding one second. The first strike following this interval is classified as a Proactive Punch, whereas strikes responding within 0.2 seconds of an opponent’s initiation are Counter Punches. Finally, Combination Complexity classifies punch sequences into linear straight combinations, rotational hook combinations, and vertical uppercut combinations based on their mechanical variety.

\subsection{Curation of Large-scale Boxing Dataset}
% Figure 2
\begin{figure}[htbp]
    \centering
    \includegraphics[width=\textwidth]{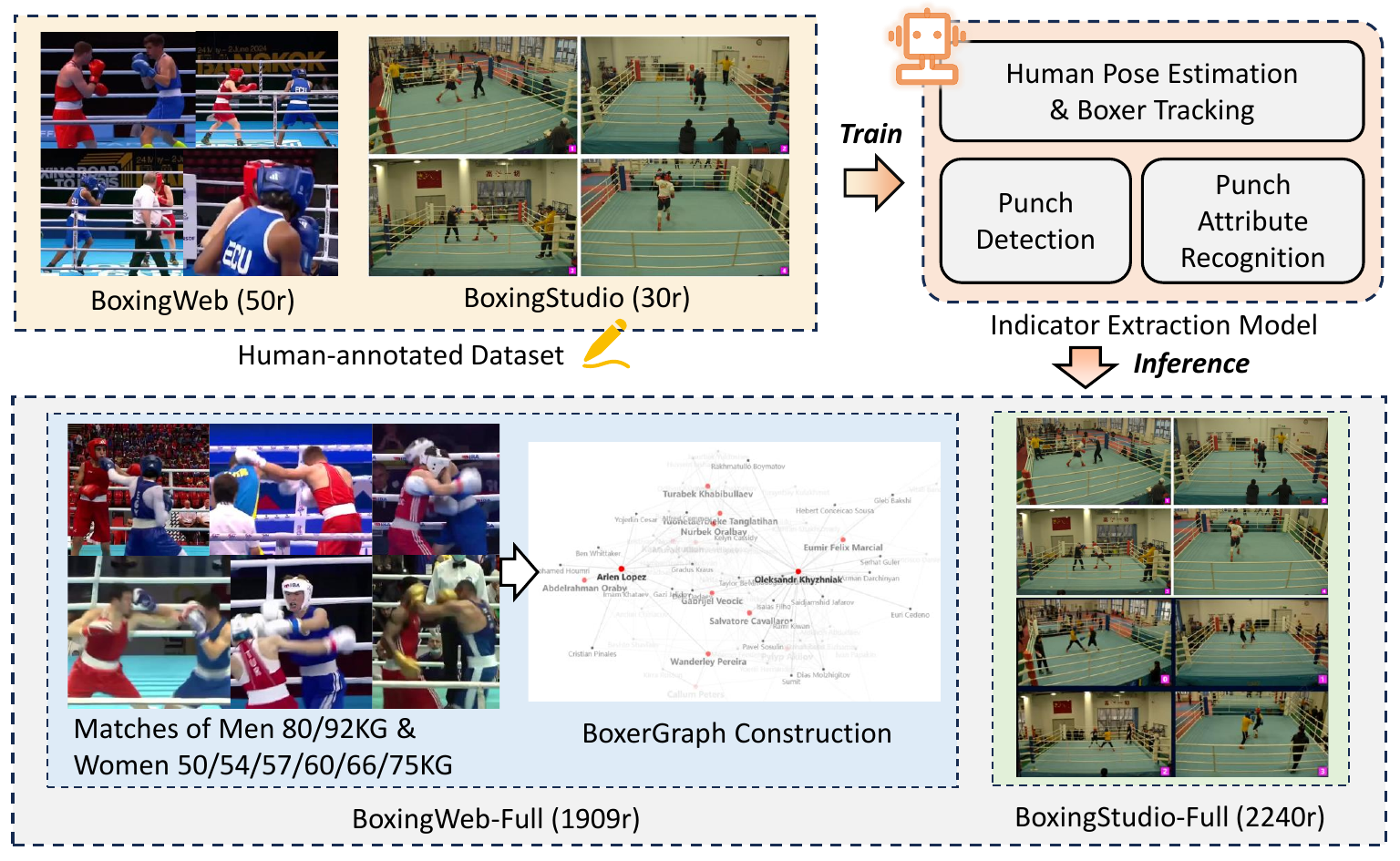}
    \caption{Data curation process for constructing the large-scale boxing dataset.}
    \label{fig:dataset}
\end{figure}
To create a robust data foundation for our analyses, we curate a multi-source, large-scale boxing dataset through a multi-stage process (Fig. \ref{fig:dataset}). The process begins with the development of two manually annotated datasets, BoxingWeb and BoxingStudio, which comprise 80 match rounds (240 minutes) and 10.9K atomic events. These human-annotated data are utilized for training and validating our vision models. 

\begin{figure}[htbp]
    \centering
    \includegraphics[width=\textwidth]{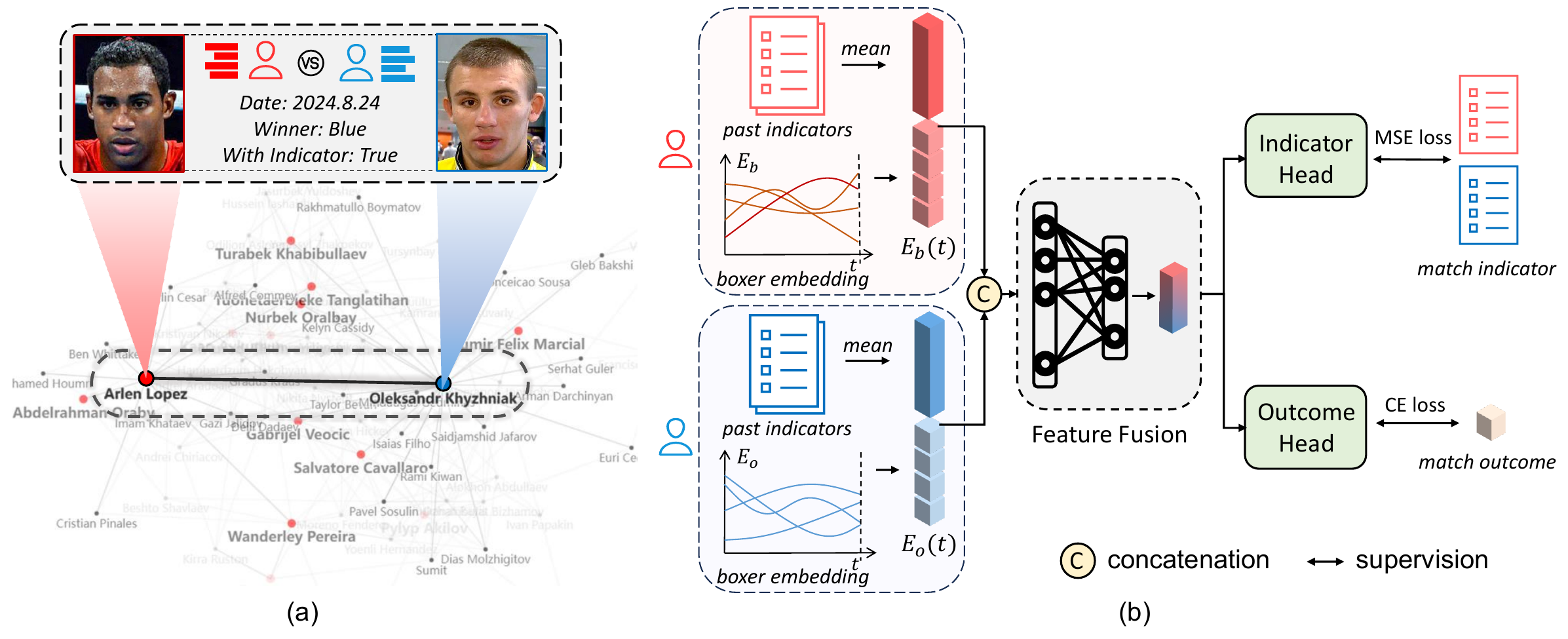}
    \caption{(a) An example of BoxerGraph (Men 80KG). (b) Architecture of the match outcome prediction model.}
    \label{fig:boxergraph}
\end{figure}

Leveraging these validated models, we process a collection of 651 publicly available matches to create the BoxingWeb-Full dataset. This dataset spans 1,909 rounds (119 hours of video) from 2021 to 2024, covering male and female athletes across eight weight classes. For each weight level, we construct a BoxerGraph (Fig. \ref{fig:boxergraph}(a)). In this topology, each boxer constitutes a node, which encapsulates both their implicit learnable embedding and their explicit historical indicator profile. An edge is established between two nodes if the boxers have competed, storing the ground-truth match outcome and the specific technical-tactical indicators generated during that bout. For the specific experiments in match outcome prediction, we utilize the BoxerGraph-80KG dataset which comprises 298 historical matches (131 with match footage) among 68 elite boxers in the 80kg weight class, enabling a focused evaluation of the predictive model. 

% Figure 3 Placeholder

In parallel with these efforts, our collaboration with the Chinese National Boxing Team involves collecting and analyzing 2,240 rounds of their training footage to monitor athlete development and inform coaches of further preparation strategies. Together, these datasets provide the foundation for both model development and practical application. 

\subsection{Graph-based Match Outcome Prediction Model}
To provide a robust foundation for strategy recommendation, we construct a unified BoxerGraph outcome prediction model that captures the dynamic interplay between observable technical styles and latent competitive hierarchies. (Fig. \ref{fig:boxergraph}(b)). 

For a match between a boxer $b$ and an opponent $o$ at time $t$, the input representation integrates two distinct components. The first is the explicit indicator profile, calculated as the average of the 18 technical-tactical indicators from the boxer's past matches to serve as the style representation ($I_{ind}$). The second is an implicit latent ability, assigned as a learnable embedding $E_b \in \mathbb{R}^{D \times C}$ to each boxer in the graph. Derived from the topology of match outcomes, this embedding captures the boxer's latent standing within the global competitive network, acting as a calibrator for the explicit indicators. To capture evolution, the embedding at time $t$ is modeled as a polynomial function of time (Eq. \ref{eq:embedding}), where $D$ is the latent feature dimension and $C$ is the time order:
\begin{equation}
    E_b(t) = \sum_{c=0}^{C-1} E_b^{(c)} \cdot t^c
    \label{eq:embedding}
\end{equation}

The model fuses these inputs to predict both the match outcome and the specific technical performance in the current match. We concatenate the past indicators and temporal embeddings of both boxers and pass them through a Multi-Layer Perceptron (MLP) \cite{rosenblatt1958perceptron} fusion module to model the interaction between the two fighters:
\begin{equation}
    F_{match} = \text{MLP}_{fusion}(I_{b,ind} \oplus E_b(t) \oplus I_{o,ind} \oplus E_o(t))
    \label{eq:fusion}
\end{equation}
The fused feature $F_match$ then branches into two parallel components: an outcome head that estimates the winning probability $\hat{y}$ and an indicator head that simultaneously forecasts the specific technical-tactical indicators $\hat{I}_{curr}$ exhibited by both boxers during the match. 

We train the model using a multi-task loss function combining prediction accuracy and feature reconstruction. We utilize Cross-Entropy (CE) loss \cite{rubinstein1999cross} for the match outcome and Mean Squared Error (MSE) loss for the indicator prediction. The indicator supervision ensures the model understand how specific styles translate to match realities:
\begin{equation}
    L_{total} = \alpha L_{MSE}(\hat{I}_{curr}, I_{GT}) + \beta L_{CE}(\hat{y}, y_{GT})
\end{equation}

\subsection{Gradient-based Strategy Recommendation}
We formulate strategy recommendation as an optimization problem aimed at identifying specific technical-tactical adjustments that maximize the winning probability against a given opponent. Since our outcome prediction model explicitly utilizes the boxer's indicator profile $I_{b,ind}$ as input, we can leverage the differentiability of the neural network \cite{rumelhart1986learning}. We compute the gradient of the predicted winning probability $\hat{y}$ with respect to the input indicators:

\begin{equation}
    G_b = \frac{\partial \hat{y}}{\partial I_{b,ind}} = \left[ \frac{\partial \hat{y}}{\partial i_1}, \dots, \frac{\partial \hat{y}}{\partial i_{18}} \right]
    \label{eq:gradient}
\end{equation}

A positive gradient $\frac{\partial \hat{y}}{\partial i_{k}}$ implies that increasing the value of indicator $k$ (e.g., increasing the proportion of lead-hand punches) will positively contribute to the winning probability against the current opponent $o$.

We rank the indicators by their gradient magnitude. The top-5 indicators with positive gradients are selected as the recommended strategic adjustments. This approach transforms the black box of the outcome prediction model into actionable, opponent-specific tactical advice.

\section{Results}

\subsection{Match Outcome Prediction Performance}
We evaluate the model on the BoxerGraph-80KG dataset (matches post-July 1, 2023). We compare our unified approach against traditional scalar rating systems and conduct ablation studies to validate the contribution of each model component.
% Table 2 
\begin{table}[htbp]
    \centering
    \caption{Accuracy comparison of match outcome prediction models.}
    \footnotesize
    \label{tab:prediction_results}
    \begin{tabular}{llcc}
        \toprule
        \textbf{Category} & \textbf{Method} & \textbf{Test Set Accuracy} & \textbf{Olympics Accuracy} \\
        \midrule
        \multirow{3}{*}{Scalar Rating Baselines} & Glicko \cite{glickman1995glicko} & 58.7\% & 75.0\% \\
         & Elo \cite{elo1967proposed} & 60.3\% & 75.0\% \\
         & WHR \cite{coulom2008whole} & 60.3\% & 75.0\% \\
        \midrule
        \multirow{2}{*}{Ablation Studies} & Explicit Indicator Profile Only & 54.0\% & 68.8\% \\
         & Latent Embeddings Only & 63.5\% & 75.0\% \\
        \midrule
        \textbf{Proposed Method} & \textbf{BoxMind (Unified)} & \textbf{69.8\%} & \textbf{87.5\%} \\
        \bottomrule
    \end{tabular}
\end{table}

As presented in Table \ref{tab:prediction_results}, our method significantly outperforms traditional rating algorithms (Glicko, Elo, WHR), which plateau at 60.3\% accuracy. This confirms that representing a boxer with a single scalar value fails to capture the multidimensional stylistic interactions inherent in elite combat sports. Ablation studies further clarify the distinct roles of the two input components. The model relying solely on explicit indicator profiles achieves only 54.0\% accuracy. This performance limitation suggests that raw statistical outputs lack competitive context: without accounting for the quality of opposition, the model cannot distinguish between high performance against weaker opponents and actual dominance against elite rivals. In contrast, the model using only latent embeddings achieves 63.5\% accuracy. This demonstrates that the learnable embeddings successfully encode the global competitive hierarchy from the match topology. By capturing the latent relative standing of athletes, this component addresses the fundamental strength differences that dominate most match outcomes.

By fusing these components, the unified BoxMind model achieves a remarkable 69.8\% accuracy on the test set, marking a 9.5\% improvement over the best baseline. This validates our hierarchical fusion hypothesis: the latent embeddings act as a competitive calibrator, allowing the model to correctly contextualize the explicit technical-tactical profiles. The model effectively learns to rely on global strength differences when the gap is significant, while pivoting to specific stylistic interactions when opponents are matched in strength.

\subsection{Strategy Recommendation Validation}
We validate the system's strategic capabilities through both quantitative comparison with human experts and a real-world case study of the Olympic gold medal journey.
% Figure 4
\begin{figure}[htbp]
    \centering
    \includegraphics[width=\textwidth]{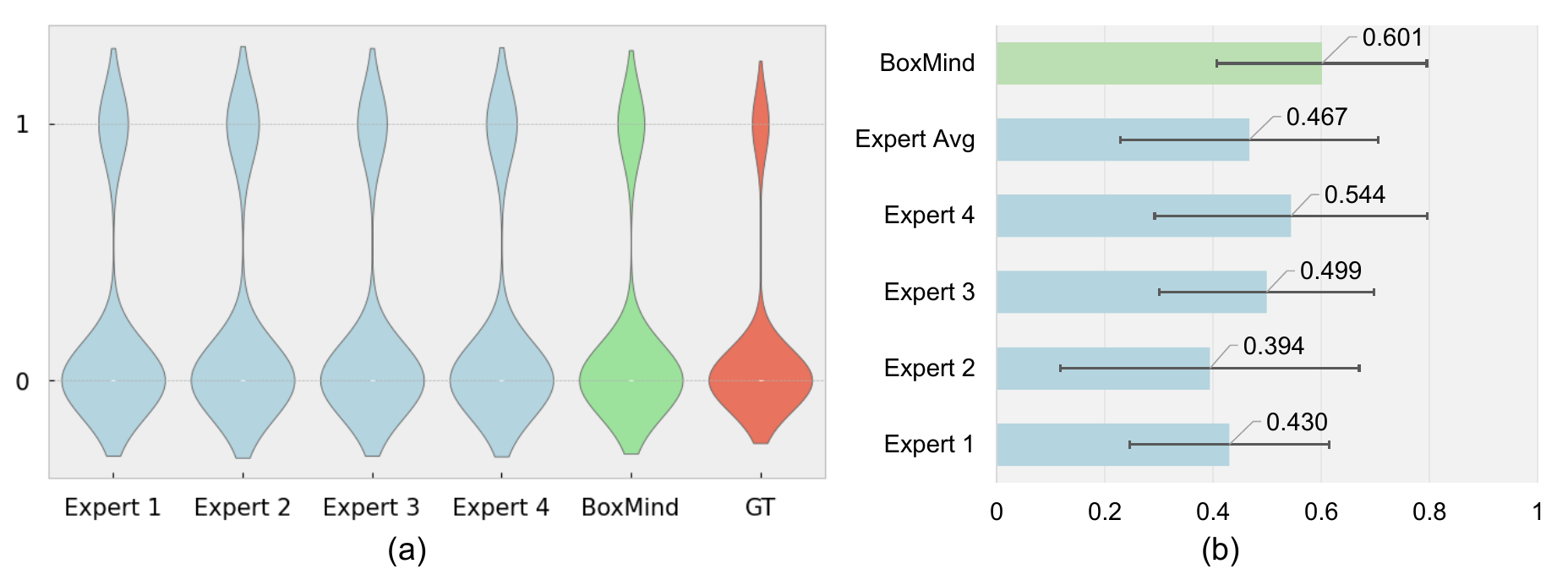}
    \caption{Comparison between BoxMind and human experts on strategy recommendation. Panel (a) shows the label distributions, while panel (b) shows the corresponding F1-scores.}
    \label{fig:expert}
\end{figure}

\textbf{Comparison with Human Experts} We evaluate BoxMind’s recommendations against those of four human experts for 10 pivotal matches from the 2024 Paris Olympics. Both the system and experts provide binary recommendations for the 18 indicators. Using the majority vote as Ground Truth, BoxMind achieves a mean F1-score of 0.601 ± 0.194, compared to the human experts' average of 0.467 ± 0.238 (Fig. \ref{fig:expert}(b)), proving it can generate professional-level tactical advice.

Statistical analysis using a paired t-test yields t = 1.623 and p = 0.111. Although the higher mean score of BoxMind does not reach statistical significance at the 0.05 level, this result establishes that BoxMind has attained professional-level proficiency comparable to human experts . Crucially, BoxMind exhibits a narrower standard deviation ($\sigma$ = 0.194 vs. 0.238), suggesting that the AI system provides more consistent and standardized tactical recommendations, effectively mitigating the subjective variance and specialization bias often observed in human analysis.

% Figure 5
\begin{figure}[htbp]
    \centering
    \includegraphics[width=\textwidth]{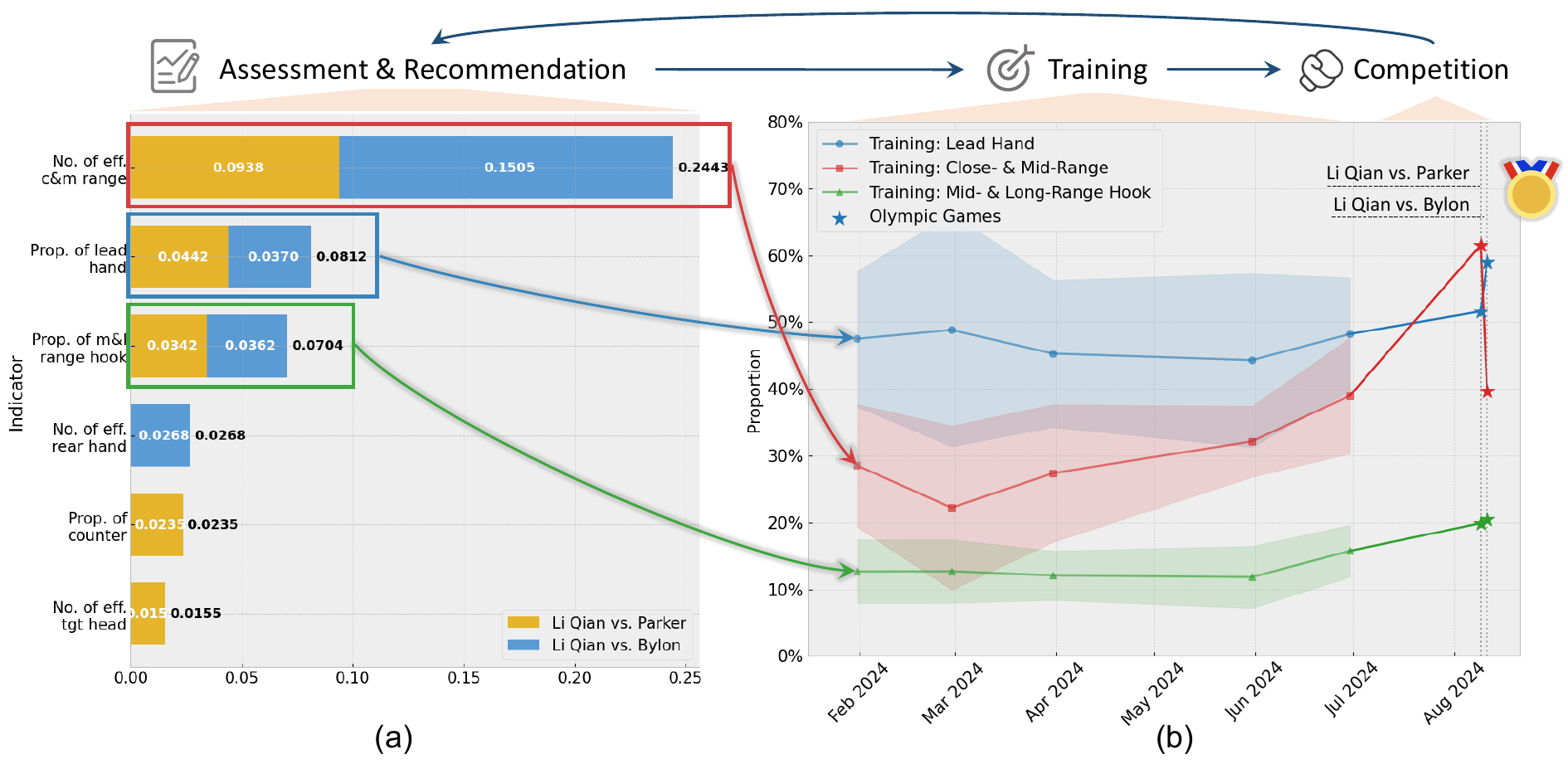}
    \caption{Case study of the Closed-Loop strategy implementation for Olympic Champion Li Qian. (a) Gradient-based strategic analysis. (b) Temporal evolution of key indicators.}
    \label{fig:case_study}
\end{figure}

\textbf{Case Study: The Closed Loop for Olympic Gold} The practical value of BoxMind is demonstrated in the preparation of Li Qian (Women’s 75kg) for the 2024 Paris Olympics, forming a verified closed loop of Assessment → Recommendation → Training → Competition.

Prior to the Olympics, BoxMind utilizes gradient analysis to evaluate Li Qian against her primary rivals, Caitlin Parker and Atheyna Bylon. As shown in Fig. \ref{fig:case_study}(a), the system identifies a high strategic gradient for three specific indicators: Number of Effective Close- \& Mid-Range Punches, Proportion of Lead Hand Punches, and Proportion of Mid- \& Long-Range Hook Punches. This suggests that to maximize her winning probability, Li Qian needs to mainly increase her engagement in the close-to-mid distance while utilizing more lead hand strikes for pace control and employing long-range hooks for guard penetration.

Guided by these recommendations, the coaching team adjusts her training regimen from January to July 2024. Monitoring data from BoxMind (Fig. \ref{fig:case_study}(b)) reveals a clear upward trajectory in these targeted metrics. By the end of the training cycle, her Proportion of Close- \& Mid-Range Punches has surged from 28.5\% to 39.0\%. Simultaneously, her Proportion of Mid- \& Long-Range Hook Punches improves by 3.1\%, and her Proportion of Lead Hand Punches sees a steady rise of 0.7\%. 

The implementation of these data-driven training adjustments is reflected in the Olympic Semi-Finals and Finals. In both matches, Li Qian executes the recommended strategic patterns with high fidelity. Compared to her performance at the end of the training camp, her Proportion of Close- \& Mid-Range Punches increases by a further 11.6\%, effectively suppressing her opponents' long-range advantages. Additionally, her Proportion of Mid- \& Long-Range Hook Punches and Proportion of Lead Hand Punches rise by 4.5\% and 7.1\% respectively. This multi-dimensional alignment between the system's recommendations and her in-competition execution established a marked tactical advantage, supporting her successful campaign for the historic gold medal.

\section{Discussion}
\subsection{Methodology}
A fundamental challenge in sports AI is the semantic gap between low-level visual signals (pixels, skeletons) and high-level abstract reasoning (tactics, strategy). End-to-end black-box models often fail to bridge this gap, offering predictions without actionable explanations. Our work resolves this by establishing a novel hierarchical abstraction layer.

Instead of relying on generic action recognition, we define the Atomic Punch Event as the fundamental unit of combat syntax. By rigorously structuring video data into discrete events with precise spatial-temporal attributes, we transform raw visual perception into a structured knowledge representation. This granular definition allows us to aggregate atomic actions into 18 high-level technical-tactical indicators, effectively creating a computable language that aligns with expert cognition.

This hierarchical construction is the prerequisite for our differentiable strategy optimization. Instead of treating match outcomes as a black-box result of latent features, we formulate the winning probability as a differentiable function of the interpretable technical indicators conditioned on the latent embeddings. This allows us to calculate the winning gradient—identifying precisely which atomic behaviors will improve the probability of victory. Consequently, the system advances the paradigm from mere observation to causal reasoning, ensuring that strategic recommendations are not only optimal but also semantically interpretable to human coaches.

\subsection{Standardizing Expert Intelligence}
A key contribution of this study is quantifying the reliability of AI decision-making. Our analysis reveals that BoxMind operates at the same level of proficiency as human experts, as indicated by the lack of statistical difference in their performance scores (p = 0.111). However, the system’s defining advantage lies in its efficiency and consistency under temporal constraints.

In dense tournament schedules, such as the Olympics, coaching teams face immense pressure to rapidly formulate strategies for consecutive opponents within limited timeframes. Traditional manual video review is labor-intensive and time-consuming, often struggling to keep pace with the competition rhythm. In contrast, BoxMind automatically synthesizes collected data to generate opponent-specific strategies immediately. This capability significantly accelerates the preparation process compared to human analysis. Consequently, the system serves as a rapid, data-driven baseline, ensuring that tactical decisions remain objective and standardized, even under the urgency of high-intensity competition.

\subsection{Practical Validation}
The most significant validation of BoxMind is its deployment as a closed-loop system during the 2024 Paris Olympics. Unlike most sports AI research that operates in an open loop—developing models on historical data without practical intervention—our work demonstrates a complete cycle: Assessment → Recommendation → Training → Competition.

The case of Olympic Champion Li Qian serves as a proof-of-concept for this paradigm. The system identifies specific tactical adjustments, most notably the need to increase engagement in close and mid-range distance, months before the Games. The quantifiable improvement in this metrics during training (+10.5\%), followed by its intensified execution in the gold medal match (+11.6\%), provides causal evidence of the system's impact. This transition from descriptive analytics to prescriptive intervention marks a milestone for AI in sports, proving that algorithmic insights can effectively translate into competitive success.

\subsection{Extensibility and Limitations}
While validated in boxing, the framework of Atomic Event → Semantic Indicator → Matchup Modeling → Gradient Optimization possesses inherent extensibility to other adversarial domains. The core logic relies on defining discrete, semantically meaningful actions and learning a predictive manifold where outcomes can be optimized. This methodology is not limited to combat sports. It can be adapted to team sports by analyzing unit interactions or to e-sports where action granularity is high. By replacing the domain-specific atomic definitions, the underlying strategy optimization engine remains universally applicable.

Regarding future developments, we aim to advance the system from offline analysis to real-time application. Currently, BoxMind operates as a post-match or pre-match planning tool. Developing a lightweight inference engine that runs on edge devices could enable immediate tactical adjustments between rounds. This evolution from strategic planning to real-time decision support represents the next key frontier for our research.

\section{Conclusion}
This paper presents BoxMind, an AI framework designed to automate the transition from visual perception to strategic reasoning in boxing. By structuring raw video data into semantic indicators and modeling matchup dynamics, we formulate strategy recommendation as a controllable gradient optimization task. This methodology overcomes the limitations of subjective human analysis by providing consistent, quantifiable, and opponent-specific tactical guidance.
The practical utility of BoxMind is validated through its full deployment in the preparation cycle for the 2024 Paris Olympics. The system’s closed-loop interventions directly support the Chinese National Boxing Team, contributing to a historic haul of three gold and two silver medals. Beyond boxing, this work establishes a replicable paradigm for applying differentiable strategy optimization to other adversarial sports, demonstrating the potential of AI to function as an active agent in high-performance competitive training.

\bibliographystyle{unsrt} % 设定正文引用格式
% \bibliography{ref_main}   % 指向 ref_main.bib 文件 (不需要加 .bib 后缀)

\newpage
\appendix
% 重置计数器，使附录的图表显示为 Fig S1, Table S1
\setcounter{figure}{0}
\setcounter{table}{0}
\setcounter{equation}{0}
\renewcommand{\thefigure}{S\arabic{figure}}
\renewcommand{\thetable}{S\arabic{table}}
\renewcommand{\theequation}{S\arabic{equation}}

\section*{Supporting Information}

\section{Datasets and Annotation Standards}
\begin{figure}[htbp]
    \centering
    \includegraphics[width=\textwidth]{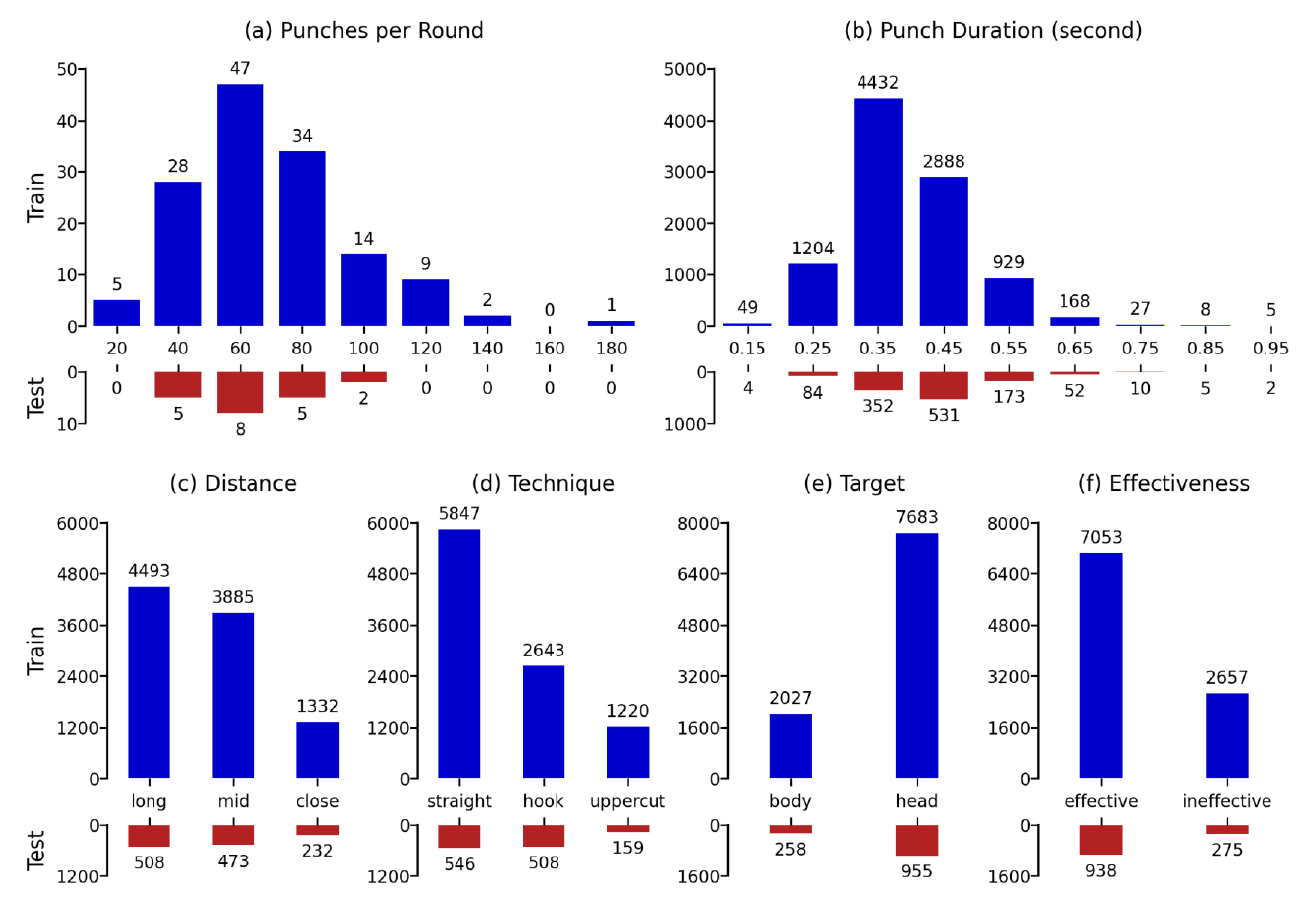}
    \caption{Distribution of the training and test sets for the indicator extraction model.}
    \label{fig:dataset_dist}
\end{figure}

To ensure both the diversity of real-world scenarios and the precision of kinematic data, we curate two complementary datasets: BoxingWeb and BoxingStudio. BoxingWeb comprises 50 rounds of high-level matches (spanning 2021–2024) collected from public broadcast sources (e.g., YouTube). This dataset captures the complexities of live competition, featuring variable resolutions ranging from 640 × 360 to 1920 × 1080 at 30 fps. BoxingStudio, collected in collaboration with the National Team using the FastMove system, consists of 30 rounds of sparring matches captured from four synchronized views at 60 fps (1920 × 1080). This dataset provides high-fidelity 3D human keypoints for training robust pose estimation. For model development, we utilize a composite training set consisting of 40 rounds from BoxingWeb and the full 30 rounds from BoxingStudio, while reserving 10 randomly selected rounds from BoxingWeb exclusively for testing. The detailed statistical distribution of the dataset across different attributes is illustrated in Fig. \ref{fig:dataset_dist}.

% Figure S1

We establish a rigorous annotation protocol to map raw footage to the structured atomic events. Professional boxing coaches manually annotate the precise temporal boundaries and attributes for every punch. Specifically, the start time ($t_{start}$) is marked at the exact moment the hand breaks from the defensive guard state to initiate a linear (Straight), downward (Uppercut), or lateral (Hook) trajectory. The end time ($t_{end}$) is marked upon the hand's complete retraction to the neutral position. 
Regarding hand usage, annotators explicitly label the physical left or right hand delivering the strike. These labels are subsequently mapped to tactical lead or rear roles based on the boxer’s stance, which is determined by the orientation of the skeletal torso plane and foot positioning (i.e., a left-foot-forward posture defines an Orthodox stance where the left hand is the Lead; conversely, a right-foot-forward posture defines a Southpaw stance where the right hand is the Lead). All other spatial and technical attributes, including distance, technique, target, and effect, are annotated according to the specific criteria (Table \ref{tab:supp_categories}).

% Table S1 
\begin{table}[htbp]
    \centering
    \caption{Categories of the four punch attributes.}
    \label{tab:supp_categories}
    \resizebox{\textwidth}{!}{
    \footnotesize
    \begin{tabular}{lp{2cm}p{10cm}}
        \toprule
        \textbf{Attribute} & \textbf{Category} & \textbf{Explanation} \\
        \midrule
        \textbf{Distance} & Long-range & Defined as a spatial interval greater than the boxer's full arm length ($d > 1.0 L_{arm}$). In this zone, the boxer cannot land a strike without advancing; they must utilize footwork to bridge the gap before delivering straight punches or hooks. \\
         & Mid-range & Defined as the interval between half and full arm length ($0.5 L_{arm} < d < 1.0 L_{arm}$). In this zone, the boxer is capable of landing effective straight punches or hooks from a stationary stance, without the necessity of preparatory footwork. \\
         & Close-range & Defined as the immediate proximity less than half an arm length ($d < 0.5 L_{arm}$). This constitutes the infighting zone where the boxer primarily relies on short-range mechanics, such as uppercuts, to attack effectively without moving their feet. \\
        \midrule
        \textbf{Technique} & Straight & A linear attack where the force is concentrated on the front of the fist. \\
         & Hook & A curved attack that moves from the side to the center, with impact concentrated on the side of the fist. \\
         & Uppercut & An upward attack where the force is focused on the top of the fist. \\
        \midrule
        \textbf{Target} & Head & A punch landed or intended to land on the head, face and neck. \\
         & Torso & A punch landed or intended to land on the torso between the neck and waist. \\
        \midrule
        \textbf{Effect} & Effective & A direct strike to the opponent’s effective areas (head, neck, and the sides of the torso above the waist) using the knuckle area of either clenched fist. \\
         & Ineffective & A strike (hitting the back of the head, back, or below the waist); strikes using the side of the glove, heel of the hand, palm, or any part of the glove other than the knuckle; merely touching the opponent without delivering any force. \\
        \bottomrule
    \end{tabular}
    }
\end{table}

\section{Technical Implementation Details}
\begin{figure}[htbp]
    \centering
    \includegraphics[width=\textwidth]{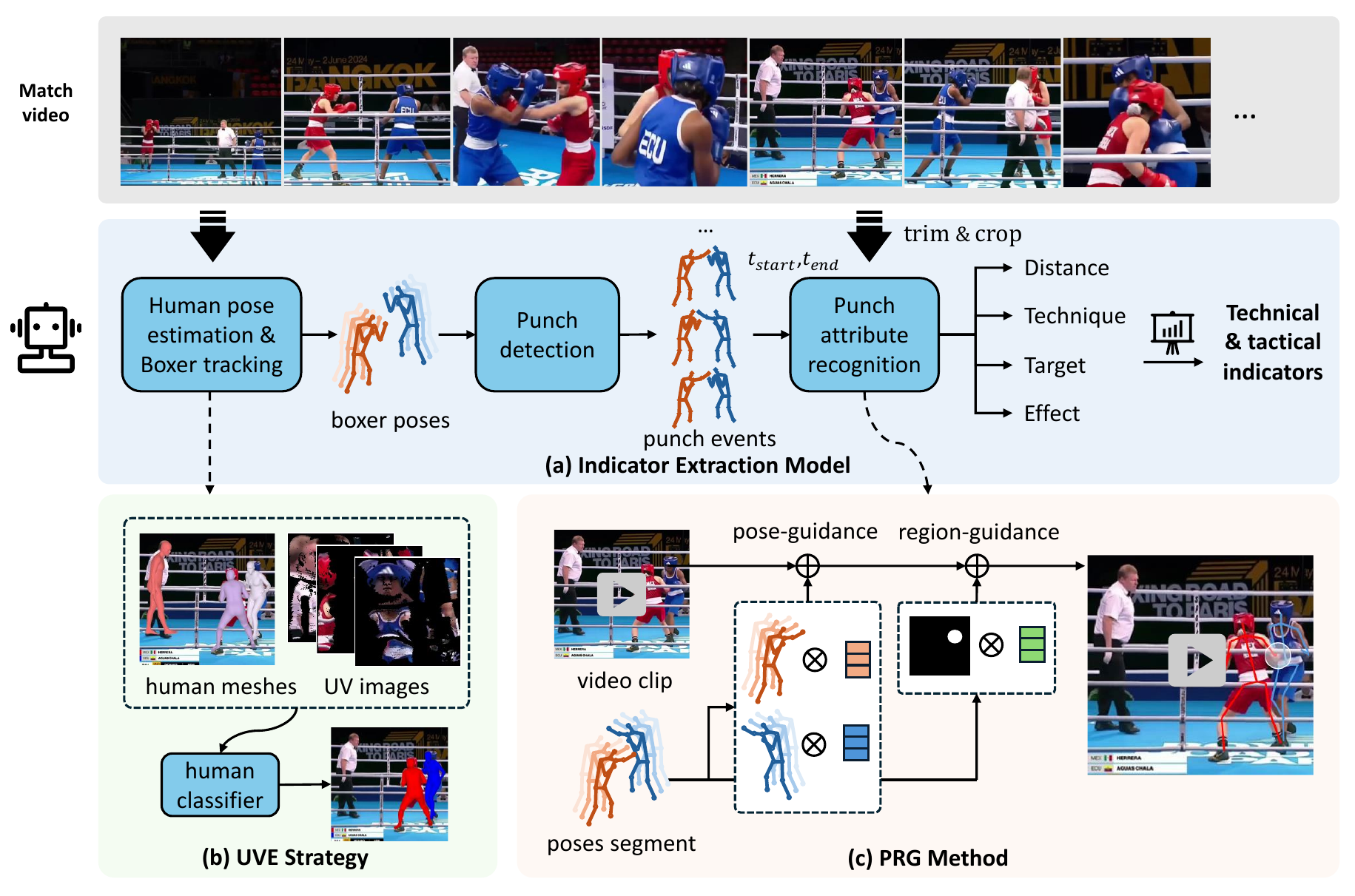}
    \caption{Framework of the automated indicator extraction model.}
    \label{fig:indicator_model}
\end{figure}

The automated extraction of technical-tactical indicators relies on a comprehensive computer vision pipeline. As illustrated in Fig. \ref{fig:indicator_model}, the framework processes raw match footage through three hierarchical modules: (1) Human Pose Estimation and Boxer Tracking, (2) Atomic Punch Event Detection, and (3) Fine-grained Punch Attribute Recognition.

% Figure S2

\subsection{Human Pose Estimation and Tracking}
The monocular videos are first fed into an off-the-shelf  human tracking and pose estimation model, 4D-Humans \cite{goel2023humans}, to estimate the SMPL \cite{loper2023smpl} parameters of all persons in the videos and to assign a unique ID to each person. For tracking, this model utilizes appearance information by extracting the UV maps of all detected persons and their location information to track them. Due to frequent shot transitions in broadcast footages and severe occlusions during close-range fighting, the tracking performance deteriorates drastically. 

To address these challenges, we propose the UV-map Enhanced (UVE) tracking strategy (Fig. \ref{fig:indicator_model}(b)), a lightweight plug-in refinement module built upon the 4D-Humans output. Every 10 frames, all tracked individuals are re-evaluated using a UV-map-based classification model \citesupp{he2016deep}, which predicts the likelihood of each person being the red-corner boxer, blue-corner boxer, or a non-boxer. If the individual currently tracked as a boxer is classified incorrectly, UVE reassigns the trajectory to the most likely candidate and updates the boxer’s identity accordingly. 

To evaluate the effectiveness of UVE, we compare its tracking performance against the original 4D-Humans on the BoxingWeb test set consisting of 10 rounds. Since 4D-Humans does not explicitly distinguish the boxers from other individuals, we manually annotate the red- and blue-corner boxers in the first frame of each video to establish a consistent identity reference for evaluation.

% Table S2 Tracking Performance 
\begin{table}[htbp]
    \centering
    \footnotesize
    \caption{Tracking performance comparison on the BoxingWeb test set.}
    \label{tab:supp_tracking}
    \begin{tabular}{lccc}
        \toprule
        \textbf{Method} & \textbf{ID Precision} & \textbf{ID Recall} & \textbf{IDF1} \\
        \midrule
        No Refinement & $0.705 \pm 0.200$ & $1.000 \pm 0.000$ & $0.793 \pm 0.168$ \\
        \textbf{UVE method} & $\mathbf{0.984 \pm 0.018}$ & $0.987 \pm 0.015$ & $\mathbf{0.985 \pm 0.013}$ \\
        \bottomrule
    \end{tabular}
\end{table}

We evaluated the strategy on the BoxingWeb test set and report the average ID precision, ID recall, and IDF1\footnote{These metrics are computed from three core quantities: identity true positives (IDTP), identity false positives (IDFP), and identity false negatives (IDFN). A prediction is counted as an IDTP when a detected person is correctly matched with the ground-truth boxer and assigned the correct identity in a given frame. If a detection is matched with a ground-truth boxer but the identity is incorrect, it is counted as an IDFP. Conversely, if a ground-truth boxer is present in the frame but not matched with any detection, it contributes to the IDFN count. Using these quantities, ID Precision is defined as IDTP / (IDTP + IDFP), reflecting how often predicted identities are correct. ID Recall is computed as IDTP / (IDTP + IDFN), measuring how well the method recovers the true boxer identities over time. Finally, IDF1 is the harmonic mean of ID Precision and ID Recall.} of the red- and blue-corner boxers to reflect how accurately and consistently each method maintains correct identity assignments over time. As shown in Table \ref{tab:supp_tracking}, the proposed UVE method significantly outperforms the baseline (No Refinement) across all identity tracking metrics. UVE achieves an IDF1 score of 0.985, compared to 0.793 for the baseline, indicating a substantial improvement in maintaining consistent boxer identities throughout the video. The improvement is primarily driven by a large increase in ID precision (0.984 vs. 0.705), reflecting UVE’s ability to suppress incorrect identity assignments caused by identity switches or tracking drift. While the baseline achieves perfect ID recall (1.000), this is expected, as the identities of the red- and blue-corner boxers are manually initialized in the first frame and remain present in all subsequent frames. However, the low ID precision and IDF1 demonstrate that without further refinement, identity consistency cannot be reliably maintained. In contrast, UVE corrects misidentifications over time, resulting in a more stable and accurate tracking of both target boxers.

\subsection{Anchor-Free Punch Detection}
To temporalize atomic actions, we design an anchor-free punch detection model. It takes the estimated 2D and 3D poses of the two boxers as input and detects the punch events for them independently. Denote the input to the detection model as $X\in R^{T\times J\times 5}$, where $T$ is the number of frames and $J$ is the number of human joints. The model first extracts the temporal information with Temporal Convolutional Networks (TCN) \cite{lin2021learning,lea2017temporal} $f_t$ by merging the joint and channel dimensions, and provides a contextualized feature for each input frame:
\begin{equation}
    F = f_t(\text{reshape}(X, [T, J \times 5])) \in \mathbb{R}^{T \times d}
\end{equation}
where $d$ is the feature dimension. The feature for each frame is then fed into two heads to detect punches of left and right hands, respectively. Each head estimates the probability that the hand is delivering a punch at the moment as well as the start and end time shift of the punch from this moment.
\begin{equation}
p_t, b_t^s, b_t^e = Head(F_t)
\end{equation}
Therefore, each detected punch can be represented by a probability, a start time, and an end time, denoted as $E_t = (p_t, t - b_t^s, t + b_t^e)$, where $t$ is the current time. Non-Maximum Suppression (NMS) \citesupp{canny1986nms} is then applied to the output for each hand to reduce duplicate detections. 
With the temporal Intersection over Union (IoU) threshold set to 0.5, the model achieves a precision of 0.806, a recall of 0.763, and an F1-score of 0.783 on the BoxingWeb test set.

\subsection{Punch Attribute Recognition}
This module classifies the fine-grained attributes of each detected punch: Distance, Technique, Target, and Effect (definitions provided in Table \ref{tab:supp_categories}). We design a two-stream model and a \textbf{P}ose-\textbf{R}egion \textbf{G}uidance (PRG) method (Fig. \ref{fig:indicator_model}(b)) to integrate the skeletal and original video data for attributes recognition. Similar to the punch detection model, the skeletal stream utilizes the TCN as the skeleton encoder to process the extracted 2D and 3D poses X. For the video stream, it leverages a pretrained Inflated 3D Convolutional Networks (I3D) \cite{carreira2017quo} model as the visual encoder to exploit the visual cues in the input RGB video $V\in R^{T\times H\times W\times 3}$. To exclude background information, the input RGB video is cropped according to the position of the concerned boxer. And to augment the visual information, one more learnable mask $V_{PRG}\in R^{T\times H\times W\times 3}$ containing the position of 2D poses and hit region is further added to the video:
\begin{align}
    F_s &= f_s(\text{reshape}(X, [T, J \times 5])) \in \mathbb{R}^{d_s} \\
    F_v &= f_v(V + V_{PRG}) \in \mathbb{R}^{d_v}
\end{align}
\noindent where $f_s$ is the skeleton encoder, $f_v$ is the video encoder, $F_s$ and $F_v$ are the extracted skeleton and video feature which have the dimension of $d_s$ and $d_v$.

There are certain correlations and conflicts between the recognition tasks of the four attributes. To better address them, we introduce the Mixture of Experts (MoE) \cite{jacobs1991adaptive} strategy into the design of the classification head. Specifically, we set $H$ experts $E_1, \dots, E_H$ and a router $R$, where each expert is a two-layer MLP and the router is a one-layer MLP. The router predicts the weight $w \in \mathbb{R}^{H \times 4}$ for each expert for each attribute. The decisions of each expert of each attribute are aggregated to come to the final recognition result:

\begin{equation}
    w = \sigma(R(F_v \| F_s))
    \label{eq:moe_weight}
\end{equation}

\begin{equation}
    y_i = \sum_{k=1}^{H} w_k^{(i)} \cdot E_k^{(i)}(F_v \| F_s), \quad i=1,2,3,4
    \label{eq:moe_output}
\end{equation}

\noindent where $\|$ denotes concatenation operation, and $\sigma$ is the softmax function.

% Table S3 Attribute Recognition 
\begin{table}[htbp]
    \centering
    \caption{Punch attribute recognition results (F1-scores).}
    \label{tab:supp_ablation}
    \begin{tabular}{lccccc}
        \toprule
        \textbf{Model} & \textbf{Distance} & \textbf{Technique} & \textbf{Target} & \textbf{Effect} & \textbf{Avg.} \\
        \midrule
        I3D (video-only) & 0.662 & 0.605 & 0.662 & 0.620 & 0.637 \\
        TCN (pose-only) & 0.664 & 0.692 & 0.747 & 0.621 & 0.681 \\
        I3D+TCN (Baseline) & 0.661 & 0.691 & 0.758 & 0.635 & 0.686 \\
        Baseline+PRG & 0.654 & 0.709 & \textbf{0.768} & 0.636 & 0.692 \\
        \textbf{Baseline+PRG+MoE} & \textbf{0.665} & \textbf{0.725} & 0.764 & \textbf{0.645} & \textbf{0.700} \\
        \bottomrule
    \end{tabular}
\end{table}

Table \ref{tab:supp_ablation} presents the comprehensive ablation study results. Prior single-modality approaches prove inadequate. The video-only I3D baseline captures contextual spatial features but lacks explicit boxer ordering, leading to the misattribution of punches during rapid exchanges. This deficiency is evidenced by significantly lower F1-scores in technique (-0.086) and target (-0.096) recognition compared to multimodal baselines. Conversely, the pose-only TCN tracks skeletal kinematics but fails to localize the subtle contact patterns critical for target and effectiveness classification, resulting in a 0.011 deficit in target recognition and a 0.014 deficit in effectiveness recognition relative to the multimodal baseline.

The baseline I3D+TCN fusion partially addresses these limitations through simple feature concatenation, achieving an average F1-score of 0.686 by combining I3D’s visual localization with TCN’s motion sequencing. However, this naive fusion underutilizes cross-modal relationships, as the temporal alignment between a boxer’s punching motion (pose stream) and the corresponding visual contact evidence (video stream) remains implicit. Our PRG method explicitly bridges this gap by projecting skeleton and punch regions onto the video channel in a learnable manner. This forces the model to attend to spatiotemporal correlations between body mechanics and strike outcomes. Consequently, this guidance mechanism boosts technique and target recognition by 0.018 and 0.010 F1, respectively, effectively resolving previously ambiguous cases where similar arm trajectories (e.g., hooks vs. uppercuts) could only be disambiguated through localized video evidence of punch angles.

Finally, optimization via MoE further enhances cross-task synergy, enabling state-of-the-art performance (0.700 Avg. F1). This is achieved by adaptively fusing modality-specific features tailored to each attribute’s discriminative requirements—leveraging pose dominance for technique classification while emphasizing video features for effectiveness discrimination.

\subsection{Training and Evaluation Details}
All models are built with PyTorch \citesupp{paszke2019pytorch} and all experiments are carried out on NVIDIA 4090 GPUs.
\subsubsection{Indicator Extraction Model}
The human classifier is trained using the Adam with decoupled weight decay (AdamW) optimizer\citesupp{loshchilov2017adamw} with a learning rate of 0.001 on a single GPU. The resolution of the input UV images is 256×256 and the training batch size is 128. The loss function is Focal Loss\citesupp{ross2017focal} and the focusing parameter $\gamma$ is 2. 

The punch detection model is trained with a batch size of 8 for 50 epochs on a single GPU using the Adam optimizer \citesupp{kingma2014adam}. The initial learning rate is 0.001 and the weight decay is 0.000001. We use focal loss with $\gamma$ = 2, setting the positive sample weight (alpha) to 0.9 and the negative sample weight to 0.1.

The punch attribute recognition model is trained using the Stochastic Gradient Descent (SGD) optimizer with an initial learning rate of 0.05, over 12 epochs on 2 GPUs. The learning rate is reduced by a factor of 10 at the 5-th and 10-th epochs. The momentum, weight decay, and batch size are set to 0.9, 0.0001, and 64, respectively. Focal Loss with $\gamma$ = 2 is used to train the model for all four attributes. For the distance attribute, the class balancing factors are set to 0.3, 0.3, and 0.4 for long-range, mid-range, and close-range punches, respectively. For the technique attribute, the balancing factors for straight, hook, and uppercut punches are 0.3, 0.3, and 0.4, respectively. For the target attribute, factors are 0.3 for head punches and 0.7 for torso punches. Finally, for the effect attribute, factors are 0.6 for effective punches and 0.4 for ineffective punches.

\subsubsection{Match Outcome Prediction Model}
For the match outcome prediction model, all samples are also used as a single batch, and the model is trained with the Adam optimizer with an initial learning rate of 0.02 for 800 epochs, applying a weight decay of 0.00001. The cross-entropy loss for win/loss prediction has a weight of 1, while the prediction error for indicator features is weighted at 0.02.

\section{Extended Experimental Results}
\subsection{Reliability and Comparisons}
To evaluate the model, we calculate the Pearson Correlation Coefficient ($r$):
\noindent The correlation is calculated as follows:

\begin{equation}
    r_k = \frac{\sum_b (i_b^{(k)} - \bar{i}^{(k)})(\tilde{i}_b^{(k)} - \bar{\tilde{i}}^{(k)})}{\sqrt{\sum_b (i_b^{(k)} - \bar{i}^{(k)})^2 \sum_b (\tilde{i}_b^{(k)} - \bar{\tilde{i}}^{(k)})^2}}, \quad 
    \bar{i}^{(k)} = \frac{\sum_b i_b^{(k)}}{\sum_b 1}, \quad \bar{\tilde{i}}^{(k)} = \frac{\sum_b \tilde{i}_b^{(k)}}{\sum_b 1}
    \label{eq:correlation}
\end{equation}

\noindent Here, $i_b^{(k)}$ and $\tilde{i}_b^{(k)}$ represent the estimated and the ground truth values of the $k$-th indicator for boxer $b$, while $\bar{i}^{(k)}$ and $\bar{\tilde{i}}^{(k)}$ are their mean values over all boxers respectively. The ground truth values of indicators are calculated based on the annotations for all test rounds.

\begin{figure}[htbp]
    \centering
    \includegraphics[width=\textwidth]{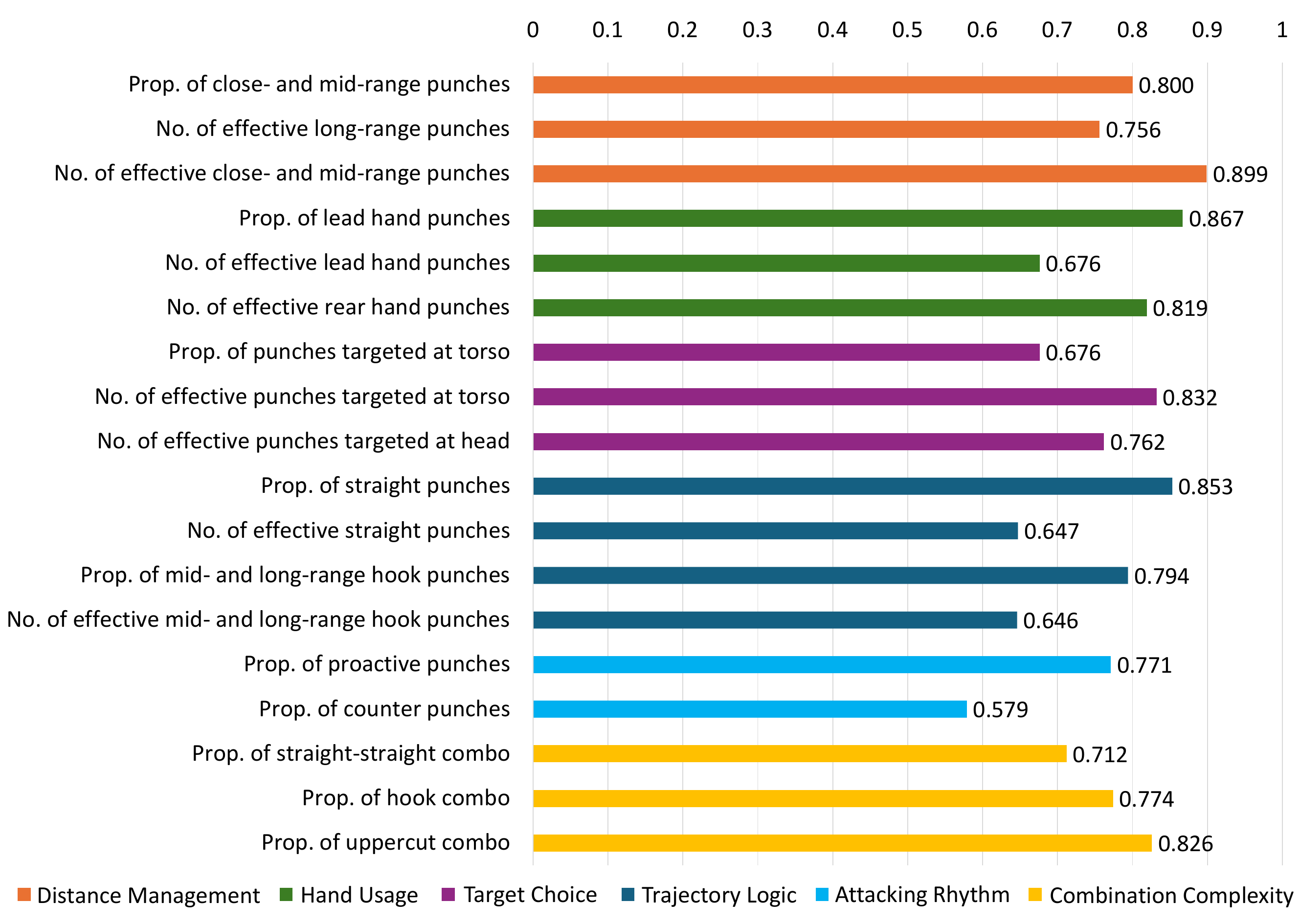}
    \caption{Correlation coefficients between estimates and ground truth for 18 indicators.}
    \label{fig:supp_corr}
\end{figure}

Fig. \ref{fig:supp_corr} visualizes the correlation coefficients across the six major indicator categories. The system achieves an overall average correlation of 0.761, proving that BoxMind can extract a boxer's stylistic signature with high fidelity. Specifically, the model demonstrates exceptional performance in Distance Management (Avg. r=0.818) and Hand Usage (Avg. r=0.787). Notably, the Number of Effective Long-Range Punches achieves the highest correlation of 0.899, and the Proportion of Lead Hand Punches reaches 0.867. This confirms that our vision pipeline (specifically the UVE tracking and PRG module) is highly robust in distinguishing fundamental spatial features, such as the spatial geometry of engagement (Long vs. Close) and the distinction between lead (control) and rear (power) hands.

In terms of Trajectory Logic, the Proportion of Straight Punches shows a high correlation of 0.853, indicating the model effectively separates linear attacks from angular ones. Furthermore, Combination Complexity maintains a strong average of 0.771, with Uppercut Combinations reaching 0.826. This suggests that the Temporal Convolutional Network (TCN) successfully aligns atomic events into coherent sequences, capturing the flow of combat.

We observe a slight performance dip in the Attacking Rhythm category (Avg. r=0.675), specifically for the Proportion of Counter Punches (r=0.579). This lower score is expected and interpretable: unlike purely visual attributes (e.g., Straight vs. Hook), defining a counter requires semantic reasoning about the interaction latency (responding within 0.2s) and intent relative to the opponent's initiation1. Detecting such subtle temporal causality is inherently more challenging than recognizing overt physical motions. However, the correlation remains positive and statistically significant, ensuring that the system can still correctly categorize a boxer as Counter-reactive versus Proactive in the macro-strategic analysis.

\subsection{Comparison with human experts on certain boxers}
\noindent To bridge the gap between quantitative metrics and qualitative expert intuition, we translate the raw indicators into semantic labels describing a boxer's specific advantages. We model the probability that a boxer $b$ outperforms the relevant population in indicator $k$ using Kernel Density Estimation (KDE). The advantage probability $p_b^{(k)}$ is formulated as the maximum likelihood of outperforming either the general capability at that weight class ($i_{\text{level}}$) or the specific historical opponents ($i_{\text{oppo}}$):

\begin{equation}
    p_b^{(k)} = \max \left( P \left( \text{KDE}_{1D}(i_b^{(k)}) > \text{KDE}_{1D}(i_{\text{level}}^{(k)}) \right), P(u > v) \right)
    \label{eq:advantage_prob}
\end{equation}

\noindent where $(u, v) \sim \text{KDE}_{2D}(i_b^{(k)}, i_{\text{oppo}}^{(k)})$. Here, $\text{KDE}_{1D}$ and $\text{KDE}_{2D}$ denote 1-dimensional and 2-dimensional kernel density estimations \citesupp{davis2011remarks,scott2011multivariate}, respectively. This formulation ensures that a boxer is credited with an advantage if they either statistically surpass the division average or consistently outperform their specific adversaries. A threshold of 0.5 is applied to map this probability into a binary advantage label (1 for Advantage, 0 for Neutral/Disadvantage).

We collect a total of 71 historical matches featuring 10 boxers who participate in the 2024 Paris Olympics and conduct the advantage analysis based on the extracted indicators from the match footage. Meanwhile, four human experts are tasked with annotating whether a boxer has an advantage in each of the 18 indicators, using a binary score of 0 or 1. To resolve inter-expert disagreement and establish a robust Consensus Ground Truth, we utilize a majority voting protocol that integrates annotations from all four experts and the BoxMind system.

\begin{figure}[htbp]
    \centering
    \includegraphics[width=\textwidth]{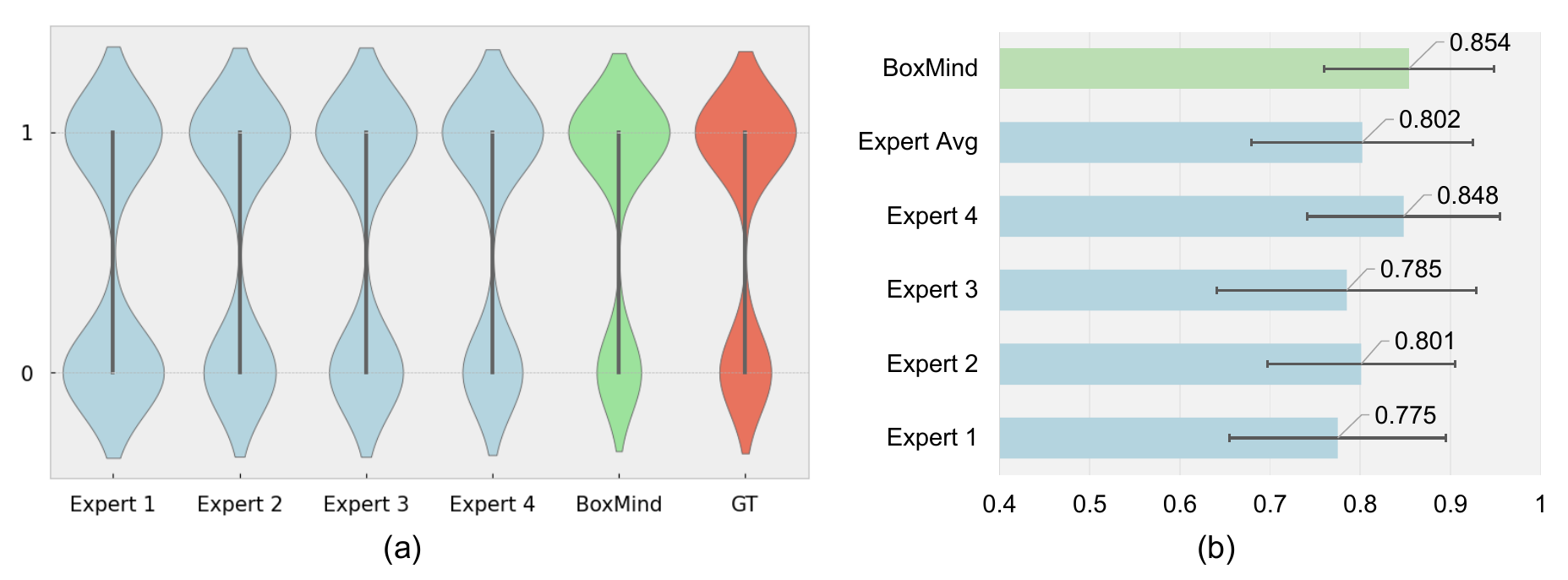}
    \caption{Results of boxer advantage analysis based on extracted indicators. (a) Label distribution for boxer advantages as independently assessed by each human expert and BoxMind. In this distribution, a label of ‘1’ indicates that the boxer is considered to have an advantage, while a label of ‘0’ indicates that the boxer is not considered to have an advantage. (b) F1-scores of boxers’ advantages assessed by each human expert and BoxMind. The majority-vote results of all experts and BoxMind are used as the ground truth (GT) for calculating the F1-scores.}
    \label{fig:supp_adv}
\end{figure}

The distribution of labels annotated by BoxMind and human experts are shown in Fig. \ref{fig:supp_adv}(a), indicating that the human experts and BoxMind approximately reach a consensus on the percentage of the boxer's advantages. Based on the voted ground truth, the F1-scores of the labels given by BoxMind and the human experts are calculated and shown in Fig. \ref{fig:supp_adv}(b). BoxMind demonstrates analytical performance comparable to that of human experts. The system achieves an F1-score of 0.854 ± 0.094, which is statistically on par with the average F1-score of the human experts (0.802 ± 0.123; t = 1.217, p = 0.230). Notably, the system's analysis exhibits lower variability ($\sigma$ = 0.094 vs. 0.123), suggesting a higher degree of objectivity and reproducibility. These findings validate that our automated pipeline not only accurately extracts performance data but also interprets it at a level equivalent to seasoned human analysts, establishing a reliable foundation for strategy recommendation.

\subsection{Case Study for Match Outcome Prediction}
\begin{figure}[htbp]
    \centering
    \includegraphics[width=\textwidth]{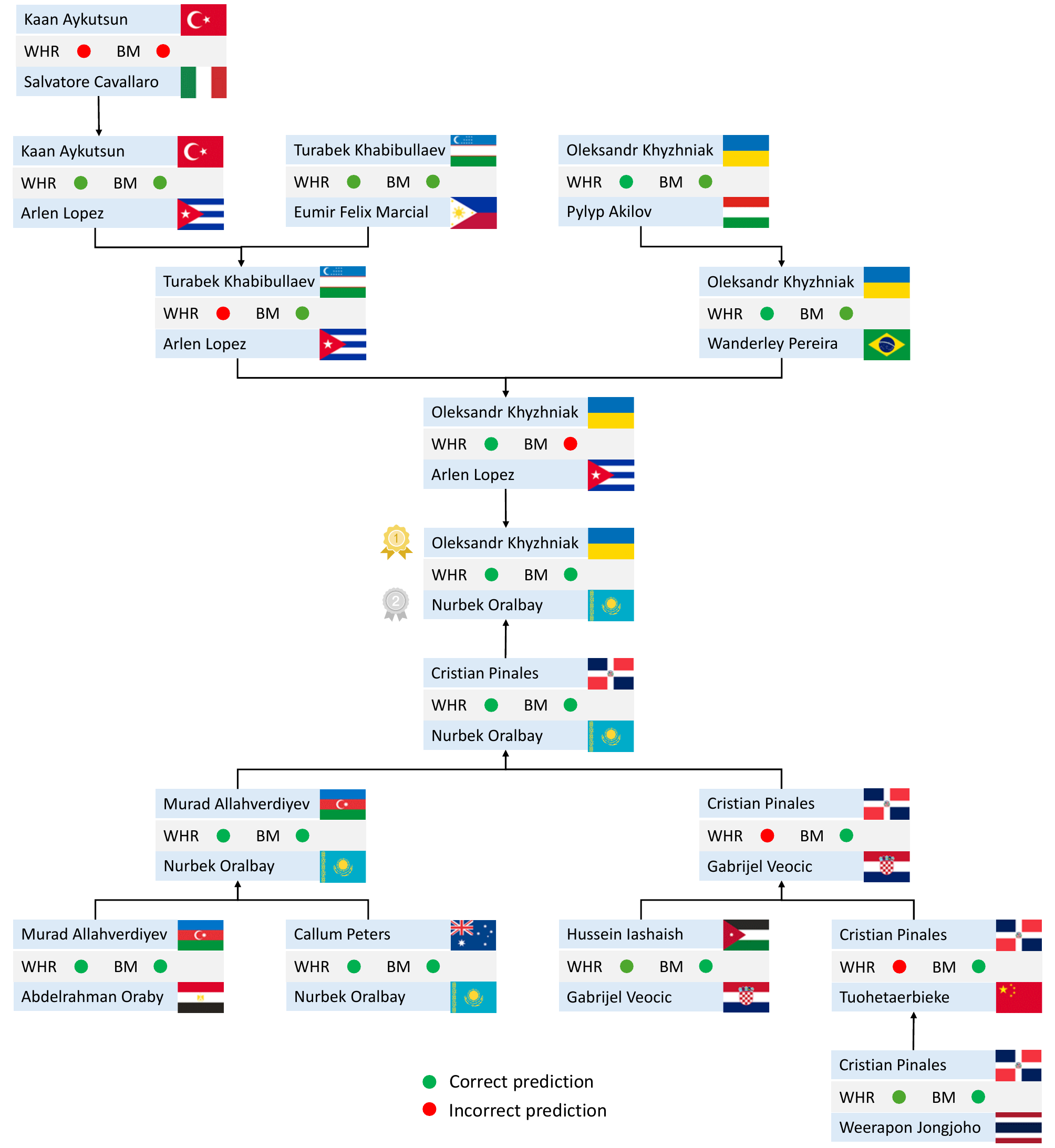}
    \caption{Match outcome prediction for the 80KG men’s boxing event at the 2024 Paris Olympics. The Whole History Rating (WHR) algorithm correctly predicts 12 out of 16 matches, while BoxMind (BM) achieves 14 correct predictions out of 16.}
    \label{fig:supp_tree}
\end{figure}

The match outcome prediction results for the 80KG men’s boxing event at the 2024 Paris Olympics is shown in Fig. \ref{fig:supp_tree}. A critical validation case is the quarter-final bout between Arlen Lopez (Cuba) and Turabek Khabibullaev (Uzbekistan). Traditional rating systems (e.g., WHR) uniformly forecast a victory for Khabibullaev, driven by his high recent activity volume and winning streak. In contrast, BoxMind correctly predicts Arlen Lopez as the victor. BoxMind identifies a critical stylistic mismatch hidden from scalar rating systems. While Khabibullaev demonstrates high Attacking Rhythm (high initiative), his extracted Combination Complexity is relatively low, indicating a reliance on linear, repetitive offense. Conversely, Arlen Lopez’s profile is characterized by high Proportion of Counter Punches and superior Spatial Control. In the actual match, Lopez effectively counters Turabek’s aggressive attacks, minimizing his own offensive risks while exploiting openings in Turabek's repeated strikes. The style dominance allows him to win the match despite not having the superior strength. This case highlights BoxMind's capacity to model non-linear stylistic compatibility, proving it to be a more nuanced predictor than history-based rating algorithms.

% 补充材料参考文献
% 注意：为了最简单的单文件编译，这里手动创建第二个参考文献列表。
% 如果需要严格的独立 BibTeX 管理，通常需要 biblatex 或 multibib 包，
% 但 arXiv 有时对这些包的处理比较复杂。手动环境是最稳妥的方法。
\renewcommand{\refname}{Supplementary References}
% 设置补充材料的引用格式 (通常和正文保持一致)
\bibliographystylesupp{unsrt} 
% 指向 ref_supp.bib 文件
% \bibliographysupp{ref_supp}

\end{document}